\documentclass{article}

    \PassOptionsToPackage{numbers, compress}{natbib}

    \usepackage[preprint]{neurips_2023}

\usepackage[utf8]{inputenc} 
\usepackage[T1]{fontenc}    
\usepackage[pagebackref=true,breaklinks=true,letterpaper=true,colorlinks,bookmarks=false]{hyperref}
\usepackage{url}            
\usepackage{booktabs}       
\usepackage{amsfonts}       
\usepackage{nicefrac}       
\usepackage{microtype}      
\usepackage{xcolor}         
\usepackage{algorithm} 
\usepackage{algorithmic}
\usepackage{amsmath}
\usepackage{graphicx}
\usepackage{adjustbox}
\usepackage{caption}

\renewcommand*\backref[1]{\ifx#1\relax \else (Cited on page #1) \fi}
\newcommand{\squeezeup}{\vspace{-2mm}}

\title{DIFF-NST: Diffusion Interleaving For deFormable Neural Style Transfer} 

\author{%
  Dan Ruta\\
  University of Surrey\\
  \And
  Gemma Canet Tarrés \\
  University of Surrey \\
  \And
  Andrew Gilbert \\
  University of Surrey \\
  \And
  Eli Shechtman \\
  Adobe \\
  \And
  Nicholas Kolkin \\
  Adobe \\
  \And
  John Collomosse \\
  University of Surrey, Adobe \\
}

\begin{document}

\maketitle

\renewcommand\twocolumn[1][]{#1}%
\maketitle
\vspace{-20pt}

    \hspace{-0.1\linewidth}\includegraphics[width=1.2\linewidth]{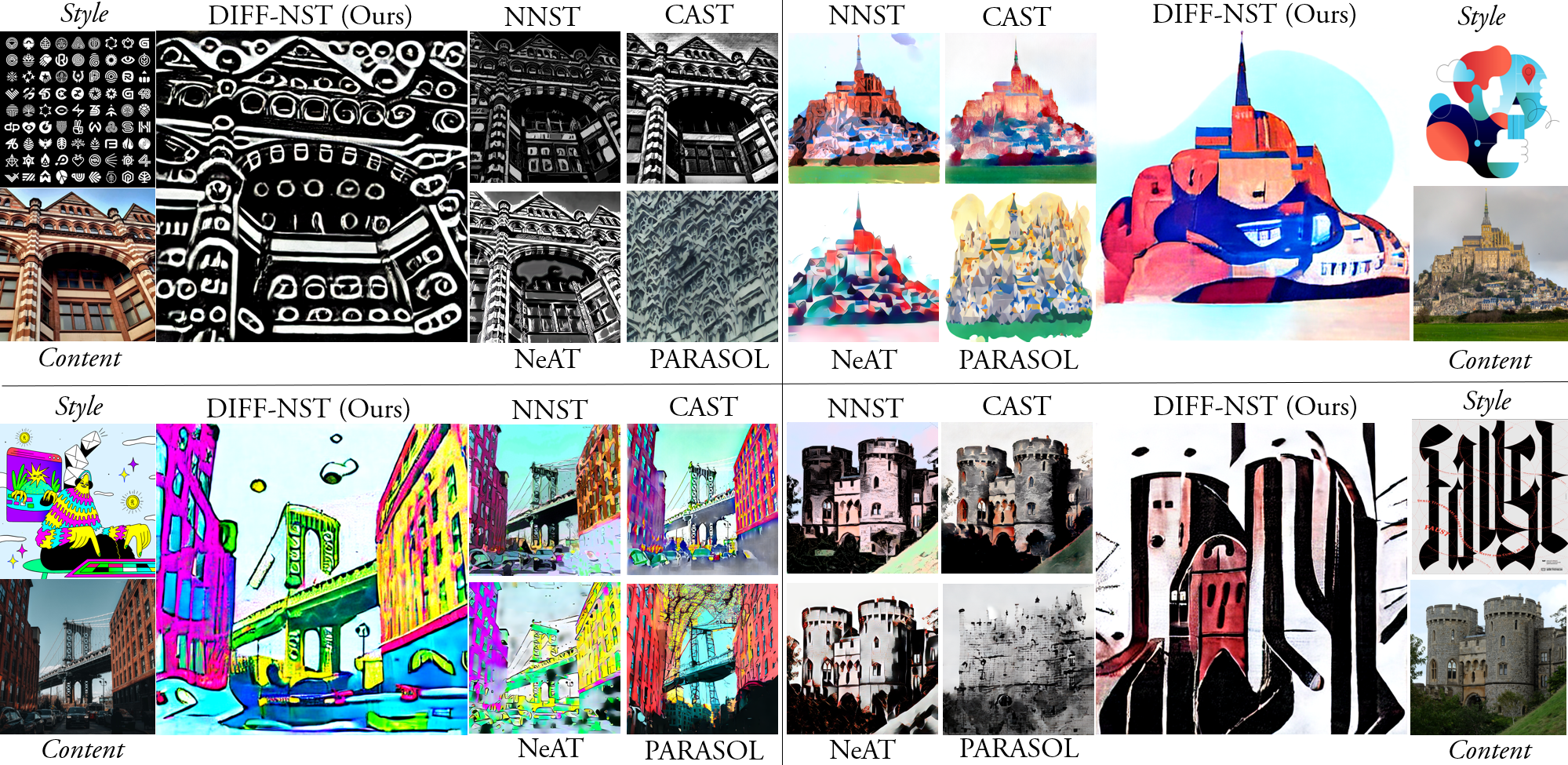}

\begin{center}
    \centering
    Figure 1. Deformable style transfer using DIFF-NST, compared to baselines: NNST \cite{nnst}, CAST \cite{cast}, NeAT \cite{neat}, and PARASOL \cite{parasol}. Our DIFF-NST method performs style transfer with much stronger style-based form alteration - matching the shapes and structures to those in the style image, not just the colors and textures. More in Fig \ref{fig:extra_deform}. Zoom for details.
\end{center}%
\vspace{9pt}

\setcounter{figure}{1}

\begin{abstract}
  Neural Style Transfer (NST) is the field of study applying neural techniques to modify the artistic appearance of a content image to match the style of a reference style image.
  Traditionally, NST methods have focused on texture-based image edits, affecting mostly low level information and keeping most image structures the same. However, style-based deformation of the content is desirable for some styles, especially in cases where the style is abstract or the primary concept of the style is in its deformed rendition of some content. 
  With the recent introduction of diffusion models, such as Stable Diffusion, we can access far more powerful image generation techniques, enabling new possibilities.
  In our work, we propose using this new class of models to perform style transfer while enabling deformable style transfer, an elusive capability in previous models. 
  We show how leveraging the priors of these models can expose new artistic controls at inference time, and we document our findings in exploring this new direction for the field of style transfer.
\end{abstract}

\section{Introduction}

Neural Style Transfer (NST) aims at re-rendering the content of one image with the distinctive visual appearance of a second style image, typically an artwork. Most prior work has focused on low level style, represented as colors and textures. However, artistic style covers a broader gamut of visual properties, including purposeful geometric alterations to the depicted content, often called \textit{form} \cite{art_form_definition}. 

We introduce a novel NST approach that considers not only low level color and texture changes but also higher level style-based geometric alterations to the depicted content. We aim to maintain the object structure to resemble the original content image and remain identifiable as such. But with style-based deformations of the content reflecting the artist's original intent as they depicted their original subject matter in the exemplar artwork image. Such content deformations have been more challenging to achieve, given a need for a higher level spatial semantic understanding of subject and/or scene information \cite{deformablenst}. 

Learning priors regarding the interplay of artistic style, semantics, and intentional deviations from photo-realistic geometry is non-trivial and not generally a part of NST pipelines. However, recent diffusion-based image generation literature has made impressive progress in modeling various visual concepts \cite{dalle2,ldm,ediffi}, accurately modeling how objects fit into the world around them.

We leverage these extensively learned priors in our work, adapting them to NST. We adapt them in our DIFF-NST model to function without text prompts in an exemplar-based setting, similar to more traditional NST. Text-less exemplar-based is desirable for some stylistic edits, as textual prompts would require extensive descriptions of the style, which may be difficult or impossible to articulate fully. We build the first NST model to make significant high level edits to content images. We compare our work to several baselines and show state-of-the-art user preference in user studies.

\section{Related Work}

The seminal work of Gatys' Neural Style Transfer (NST) \cite{gatys} enabled neural techniques for transferring the artistic style appearance of a reference artwork to an unstylized depiction of some content - typically a photograph. Follow-up works created feed-forward, optimization free approaches to achieve this \cite{HuangAdaIn2017,wct}. Other techniques for NST emerged, such as optimal transport \cite{optimaltransfernst}, hyper-networks \cite{hypernst}, and Neural Neighbours \cite{nnst}. Attention based techniques later emerged \cite{sanet,pama}, with further follow-up improvements to contrastive losses \cite{contraAST,cast}, and scaling to high resolution with improvements to robustness and detail propagation \cite{neat}. Deformation in style transfer has been explored in previous work \cite{deformablenst}, based on detecting shared keypoints between the style and content, thereby limited by a shared depicted subject. Regarding fine-grained representation space for artistic style, ALADIN \cite{aladin} introduced the first solution to this training over their fine-grained BAM-FG dataset. This was later evolved into ALADIN-ViT \cite{stylebabel} using a Vision Transformer \cite{vit} for stronger expressivity, and later as ALADIN-NST \cite{aladinnst}, with stronger disentanglement between content and style by changing BAM-FG \cite{aladin} for a fully disentangled, synthetic dataset.

Within the generative image domain, sizeable text-to-image diffusion models such as Dall-e 2 \cite{dalle2}, Parti \cite{parti}, Imagen \cite{imagen}, and e-Diffi \cite{ediffi} have recently made significant advances in image generation fidelity and control, enabling free-form text prompts as an input control vector for guiding image synthesis, with unprecedented quality. These models are trained on large datasets and require prohibitive amounts of computation. Latent Diffusion Models \cite{ldm} introduced the concept of applying the diffusion process to a smaller, latent representation of images rather than operating in pixel space like the previous works. This dramatically reduces the compute requirements for training and, more importantly, inference. Stability AI \cite{Stabilityai} democratized comprehensive open access to such models by open sourcing weights for an LDM trained on a subset of the LAION \cite{laion} dataset. 

Much follow-up research has been enabled and built on these pre-trained weights, known as the Stable Diffusion model. Due to the still prohibitive training costs, several works have studied the personalization of existing pre-trained model weights for new concepts, such as Dreambooth \cite{dreambooth}, Textual Inversion \cite{textualinversion}, and Custom Diffusion \cite{customdiffusion}. Other works have studied enabling new ways to control these models for tasks such as subject-oriented editing \cite{diffediting1,diffediting2,diffediting3}. Or focusing on more general image editing based on text-based prompt changes \cite{cyclediffusion,prompt2prompt,diffusiondisentanglement}. However, most of these techniques aim at semantic changes or require text-based prompt changes. Text-less exemplar-based stylistic edits have not commonly been explicitly explored with diffusion models. Recently, PARASOL \cite{parasol} has used an ALADIN-ViT style embedding to perform style-based image generation, with some capabilities of maintaining content structure.

\section{Method}

To push beyond the traditional boundaries of texture-only style transfer, we wish to leverage the significant learned model priors such as Stable Diffusion \cite{ldm}, having been trained on large amounts of data, with typically inaccessible amounts of compute. In our approach, as shown in Figure~\ref{fig:method} we freeze the pre-trained weights and train several modules of fully connected layers in each UNet self-attention block. We interleave pre-extracted content noise used for shapes and composition and the style attention values from the style image. These are used across reverse diffusion timesteps, generating a final stylized image using content and style information extracted from the interleaved data.

\subsection{Preliminary analysis of style information in attention space}

Prior work \cite{diffusiondisentanglement} has shown that early diffusion timesteps affect an image's global structural and compositional information, whereas later timesteps affect local fine details. Inspired by this, we set out to determine which timesteps of the diffusion process control style and which control content.

Given a lack of research around exemplar-based Neural Style Transfer with diffusion models, we use a prompt-based model, prompt-to-prompt \cite{prompt2prompt}, to carry out this visualization. We use ChatGPT \cite{chatgpt} to generate 20 content prompts, and we further define 10 style modifier prompts. With the prompt-to-prompt pipeline (operating over the Stable Diffusion LDM weights), we use the content prompts to generate reference content images, and we combine each content prompt with each style modifier prompt to re-generate the content images with the different explicitly defined styles still using prompt-to-prompt~\cite{prompt2prompt}. At the end of the process, we have 20 reference content example images and 200 "stylized" images. During the generation process, we extract attention values for analysis. We average the differences between the content example images' attention values and each of their 10 stylized variants', at each timestep. Fig 1 in the supplementary materials visualizes the average differences between these attention values at the diffusion timesteps. The red indicates a larger difference between the original content image and its stylized versions. Given that the structural and compositional information of the example content and their "stylized" counterparts is similar, we can infer that the stylistic differences relate to the higher attention discrepancies found at the later timesteps. This preliminary exploratory experiment clarifies the different effects of diffusion timesteps across the LDM generation process. 

An additional preliminary experiment using these prompt-to-prompt images is an analysis of where the style information is captured in the LDM activations. We explicitly focus on the attention mechanism, where $\mathcal{Q}$, $\mathcal{K}$, and $\mathcal{V}$ values are used in the attention process \cite{attention}. We generate a base non-stylized image with the content prompt and then stylized variants with style modifier prompts. We extract attention values from the content-only prompt generation and replace the attention values of the stylized generation with those from the content-only generation. Doing so re-generates the original, non-stylized image. However, in our analysis, we observe that interpolating between the $\mathcal{V}$ self-attention values of the content/style-modified generations (while using only the original content values for the rest) can provide control over the stylization strength. From this experiment, we can infer that most, if not all, style information is captured from just the $\mathcal{V}$ self-attention values in the LDM. We visualize examples of this style interpolation in the supplementary materials.

\subsection{DIFF-NST real image inversion}

Our work aims to perform style transfer of existing real user-provided images. As such, the re-styled synthesized image must stay faithful to the provided content image in terms of overall composition and structure. This means we must \emph{edit} the image rather than \emph{re-generate} a semantically similar approximation. We invert the content image through the LDM, similar to previous works such as prompt-to-prompt \cite{prompt2prompt} and diffusion disentanglement \cite{diffusiondisentanglement}. This inversion process extracts the predicted noise at each timestep, as predicted by the UNet modules. To reconstruct the same image using an LDM, this content noise can be injected into the reverse diffusion process, replacing the LDM noise predictions at multiple timesteps. The more timesteps the noises are applied to, the better the reconstruction fidelity, with less freedom of input from the LDM. As shown in the diffusion disentanglement work \cite{diffusiondisentanglement}, applying changes to the diffusion values from an earlier timestep allows more significant change in image structure. 

Similar to these previous works, we use 50 time steps for the forward (inversion) and reverse (re-generation) diffusion processes. However, unlike these previous works, we interleave this noise starting from an earlier time, step 5, rather than 16, to improve reconstruction quality. We apply noise until step 45 instead of 50 to allow the model to self-correct some artifacts. Also, unlike prior work, we do not set the LDM predicted noises to zero for timesteps where pre-extracted content noises are not injected into the diffusion process. We aim to allow the model to generate new details to leverage its learned priors.

A notable trait of image-to-image and image-inversion with diffusion models is that color information is not disentangled from overall image structure across timesteps, as it is with feature activation across layers of a VGG model, for example. Thus, color information must be explicitly handled before inversion. Similar to previous works \cite{art_rad_fields,neat}, we pre-adjust the color of the content image through mean and covariance matching. We do this dynamically during training before inversion.

A final consideration is that we aim to perform prompt-less execution of LDMs, given our use of exemplar images for both content and style. As such, we only need to use the model's unconditional capabilities. Latent Diffusion Models execute two iterations of their model: one with no prompt conditioning and one with prompt conditioning. The output of both branches is joined at every time step via the classifier free guidance (CFG). This exposes prompt control via this adjustable strength. Given that we aim not to use any text prompts anywhere in the process, we, therefore, altogether disable the prompt-conditioned branch of the model execution and use only the un-conditional branch for both inversion and reverse diffusion. The process would function the same if the text prompt were fixed to a generic prompt throughout or if CFG was zero, but this approach saves on compute.
    
\begin{figure}
    \centering
    \includegraphics[width=\linewidth]{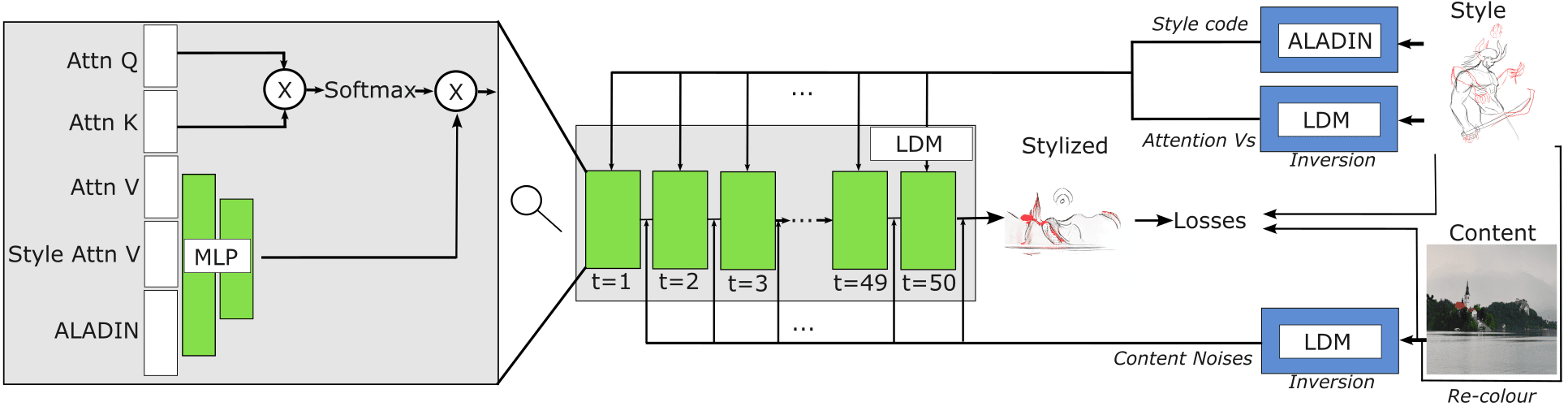}
    \squeezeup
    \squeezeup
    \squeezeup
    \caption{High level visualization of our diffusion-based neural style transfer process. (left) Trainable MLP in the self-attention blocks of the LDM Unet modules. (right) Attention values and ALADIN style codes are extracted from the style image. The content image is re-colored by the style image, after which the LDM extracts content noises from it. These are interleaved into the reverse diffusion process at multiple time steps to generate a stylized version for the loss objective. Green modules are trainable, and blue modules are frozen.}
    \squeezeup
    \squeezeup
    \label{fig:method}
\end{figure}

\subsection{Attention manipulation}

We train a set of MLPs across each self-attention module in the LDM UNet blocks. We do not wish to re-train or fine-tune the LDM weights due to large compute/financial requirements. Instead, we train several smaller modules to \emph{hijack} part of the LDM process, similar to how content noises are injected into the diffusion process. We directly target the attention process's $\mathcal{V}$ values, generating brand new values for the remaining process to use. We chose the $\mathcal{V}$ values following our initial exploratory experiments with existing text-prompt-based diffusion image editing techniques such as prompt-to-prompt, where we observed that interpolation between $\mathcal{V}$ values only is enough to induce stylistic changes between content prompts and style-modified prompts. 

Before our reverse diffusion process, similar to the real content image inversion to collect the noise predictions for reconstruction, we additionally invert and fully reconstruct the real style image through the LDM. This time, instead of collecting the predicted noises, we collect the predicted attention $\mathcal{V}$ values at every location and timestep and interleave them into the reverse diffusion process. Here, the MLPs generate the new $\mathcal{V}$ values based on an input consisting of the current $\mathcal{V}$ values, the corresponding $\mathcal{V}$ values at the same location and timestep of the style image, and the ALADIN style code of the style image, which we also pre-extract. We use both the style attention values and ALADIN, as this provides both global and local style information. Using only the attention values induces a similar style transfer. Anecdotally, however, using both sources of style information leads to a higher overall perceived quality of style transfer. We use the more recent ALADIN-NST \cite{aladinnst} variant of ALADIN, as it is more disentangled, capturing less content information. This helps to avoid semantic content creeping into the stylized image from the style image, as shown in Fig \ref{fig:horror_faces}. 

\begin{figure}
    \centering
    \includegraphics[width=0.6\linewidth]{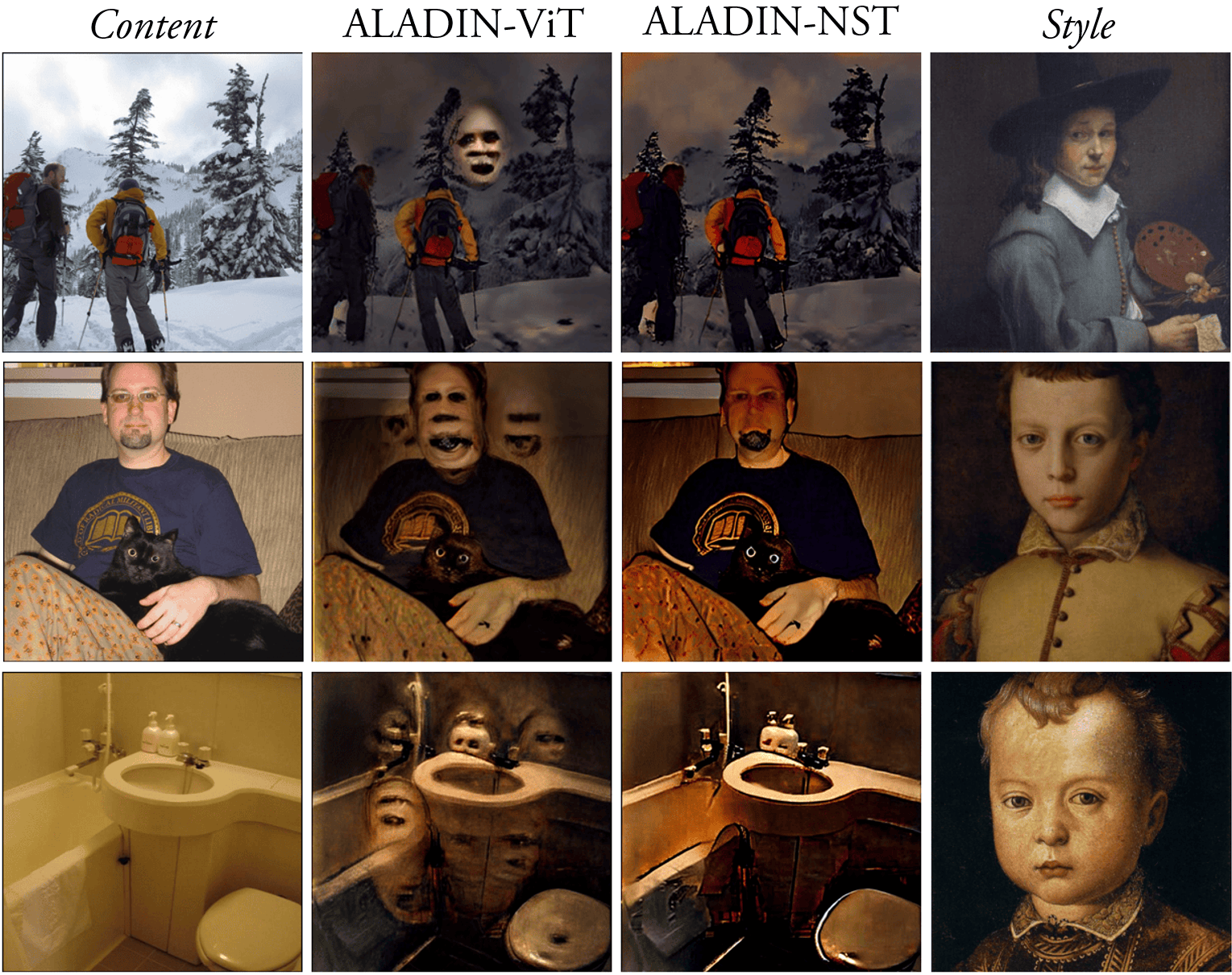}
    \caption{Visualization of style code ablation. The more disentangled ALADIN-NST \cite{aladinnst} embedding carries over less semantic information from the style images.}
    \label{fig:horror_faces}
    \squeezeup
    \squeezeup
\end{figure}

A final consideration is that we only apply this attention manipulation process to the UNet decoder/upscaling layers, as per ControlNet \cite{controlnet}. Similar to their findings, we notice no perceivable differences in the output quality, but the VRAM consumption and compute costs are lower.

\subsection{Training process}

Diffusion models are typically trained one random timestep at a time, given the nature of focusing the training on noise predictions at individual timesteps. In our case, however, such timestep-localized deltas are not as easy to isolate. We can only guide our model during training based on the final de-noised output image. Moreover, well known existing style losses have been designed to operate in pixel space. They are, therefore, not directly applicable to latent space - though this may be an area of potential future study.

Therefore, we build our training process around unrolling the entire diffusion process, from starting to ending timesteps. We then decode the latent values into pixel space, where we can finally apply standard NST losses amongst the stylized and real style images from our style dataset. We opt to keep these style learning losses similar to previous works to reduce variables and uncertainty from our work. We follow a similar training objective to recent works such as NeAT \cite{neat}, ContraAST \cite{contraAST}, and CAST \cite{cast} - described in detail in Sec \ref{sec:training_obj}. We can report some negative results in using the LDM UNet as a noised feature extractor for computing a VGG-like style loss to avoid the unrolling process - the features extracted by the UNet did not accurately model the image style features.

\subsection{Training objective}
\label{sec:training_obj}

We train our model using well explored training objectives from traditional NST methods to focus solely on the model technique - we most similarly follow training objectives resembling those of NeAT \cite{neat}, ContraAST \cite{contraAST}, and CAST \cite{cast}. Between style and stylized images, we use a VGG \cite{vgg} style loss (Eq. \ref{eq:Ls}), identity loss (Eq. \ref{eq:Lidentity_style}), contrastive loss (Eq. \ref{eq:LcontraS}), sobel-guided patch discriminator (Eq. \ref{eq:Lpatch}), domain-level discriminator (Eq. \ref{eq:Ladv}), and ALADIN loss (Eq. \ref{eq:Laladin}). Between the stylized and content images, we use a perceptual loss (Eq. \ref{eq:Lpercep}), contrastive loss (Eq. \ref{eq:LcontraC}), and identity loss (Eq. \ref{eq:Lidentity_content}). We use Sobel guidance for the patch discriminator, as per NeAT. 

Equation \ref{eq:Ls} shows the VGG style loss, with $\mu$ and $\sigma$ representing the mean and standard deviation of extracted feature maps, $I_s$ represents style image from the style dataset $S$, $I_c$ represents a content image from the content dataset $C$ after the color adjustments, and $I_{sc}$ represents the stylized image.

\begin{equation}
    \mathcal{L}_s := \lambda_{\text {vgg}}\left(  \sum_{i=1}^L\left\|\mu\left(\phi_i\left(I_{s c}\right)\right)-\mu\left(\phi_i\left(I_s\right)\right)\right\|_2+\left\|\sigma\left(\phi_i\left(I_{s c}\right)\right)-     
    \sigma\left(\phi_i\left(I_s\right)\right)\right\|_2  \right)
    \label{eq:Ls}
\end{equation}

Eq \ref{eq:Ladv} represents the domain-level adversarial loss, as per ContraAST \cite{contraAST}, learning to discriminate between generated stylized images and real artworks. Here, a discriminator $\mathcal{D}$ operates over the stylized image, following our model $M$  modules. Eq \ref{eq:Lpercep} details standard perceptual loss, where $\phi_i$ represents the pre-trained VGG-19 layer index.
\squeezeup
\squeezeup

\begin{equation}
    \mathcal{L}_{a d v} := \lambda_{\text {adv}}\left(  \underset{I_s \sim S}{\mathbb{E}}\left[\log \left(\mathcal{D}\left(I_s\right)\right)\right]+ \underset{I_c \sim C, I_s \sim S}{\mathbb{E}}\left[\log \left(1-\mathcal{D}\left( M \left( I_{s}, I_{c} \right)   \right)\right)\right] \right)
    \label{eq:Ladv}
\end{equation}

\begin{equation}
    \mathcal{L}_{\text{percep}} := \lambda_{\text {percep}}\left( \left\|\phi_{\text {conv4\_2 } }\left(I_{s c}\right)-\phi_{\text {conv4\_2}}\left(I_c\right)\right\|_2 \right)
    \label{eq:Lpercep}
\end{equation}

Eqs \ref{eq:Lidentity_style} and \ref{eq:Lidentity_content} show MSE identity losses between the reconstructed images and the style or content images, respectively. Eq \ref{eq:Laladin} shows the ALADIN loss, with $\mathcal{A}$ representing the ALADIN model. 
\begin{equation}
    \mathcal{L}_{\text {id\_s }} := \lambda_{\text {identity}}\left(\left\|I_{s s}-I_s\right\|_2\right)
    \label{eq:Lidentity_style}
\end{equation}

\begin{equation}
    \mathcal{L}_{\text {id\_c }}:=\lambda_{\text {identity}}\left(\left\|I_{c c}-I_c\right\|_2\right)
    \label{eq:Lidentity_content}
\end{equation}

\begin{equation}
    \mathcal{L}_{\text {aladin}}:=\lambda_{\text {aladin}}\left(\left\| \mathcal{A}(I_{s c}) - \mathcal{A}(I_{s}) \right\|_2\right)
    \label{eq:Laladin}
\end{equation}

Eqs \ref{eq:LcontraS} and \ref{eq:LcontraC} show contrastive losses as detailed in Sec 4.1, similar to \cite{contraAST} and \cite{cast}, where $l_s$ and $l_c$ are extracted style/content embeddings respectively, using a projection head, and $\tau$ is the temperature hyper-parameter. The contrastive losses are applied over the averaged attention values per timestep.
\begin{equation}
    \mathcal{L}_{s\text {\_contra }} := \lambda_{\text {c}}\left(  -\log \left(\frac{\exp \left(l_s\left(s_i c_j\right)^T l_s\left(s_i c_x\right) / \tau\right)}{\exp \left(l_s\left(s_i c_j\right)^T l_s\left(s_i c_x\right) / \tau\right)+\sum \exp \left(l_s\left(s_i c_j\right)^T l_s\left(s_m c_n\right) / \tau\right)}\right) \right)
    \label{eq:LcontraS}
\end{equation}
\begin{equation}
    \mathcal{L}_{c\text {\_contra }} := \lambda_{\text {c}}\left( -\log \left(\frac{\exp \left(l_c\left(s_i c_j\right)^T l_c\left(s_y c_j\right) / \tau\right)}{\exp \left(l_c\left(s_i c_j\right)^T l_c\left(s_y c_j\right) / \tau\right)+\sum \exp \left(l_c\left(s_i c_j\right)^T l_c\left(s_m c_n\right) / \tau\right)}\right) \right)
    \label{eq:LcontraC}
\end{equation}

The $\mathcal{L}_{\text {p }}$ term defined in Eq \ref{eq:Lpatch} is our patch discriminator $D_{\text {patch }}$ loss, guided by Sobel Maps ($SM$).
\begin{equation}
    \mathcal{L}_{\text {p }} = \lambda_{\text {patch}}\left( \underset{I_s \sim S}{\mathbb{E}}[-\log (D_{\text {patch }} (\operatorname{crop} ( I_{s c}, SM_{s c} ), \operatorname{crops}(I_{s}, SM_{s} ) ) ) ] \right)
    \label{eq:Lpatch}
\end{equation}

Our final combined loss objective is shown in \ref{eq:Lfinal} where each term is weighted by their respective $\lambda$ term. The loss weights are as follows: $\lambda_{\text {vgg}} = 0.5$, $\lambda_{\text {adv}} = 5$, $\lambda_{\text {percep}} = 6$, $\lambda_{\text {identity}} = 100$, $\lambda_{\text{aladin}} = 10$, $\lambda_{\text{c}} = 1$, $\lambda_{\text{patch}} = 10$, $\lambda_{\text{1}} = 0.25$, $\lambda_{\text{2}} = 0.75$.

\squeezeup
\begin{equation}
    \mathcal{L}_{\text {final }}:= \mathcal{L}_s + \mathcal{L}_{\text{adv}} + \mathcal{L}_{\text{percep}} + \mathcal{L}_{\text {id\_s}} + \mathcal{L}_{\text {id\_c}} + \mathcal{L}_{\text {aladin}} + \mathcal{L}_{\text {s\_contra }} + \mathcal{L}_{\text {c\_contra }} + \lambda_1 \mathcal{L}_{\text {p\_simple}} + \lambda_2 \mathcal{L}_{\text {p\_complex}}
    \label{eq:Lfinal}
\end{equation}
\squeezeup

\begin{figure}
    \centering
    \includegraphics[width=0.8\linewidth]{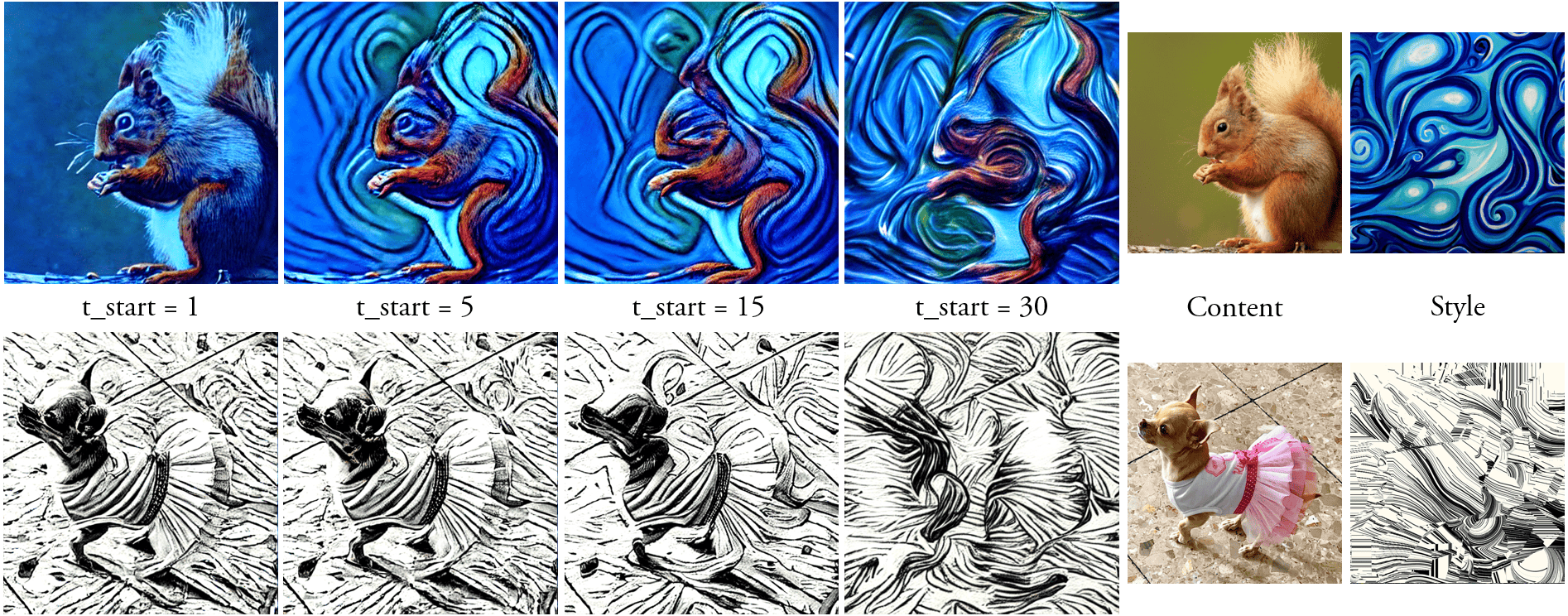}
    \squeezeup
    \caption{Controlling the style-based content deformation of the stylized image at inference time by varying the starting timestep to apply pre-extracted content noises from the content image inversion.}
    \label{fig:starting_noises}

    \includegraphics[width=0.8\linewidth]{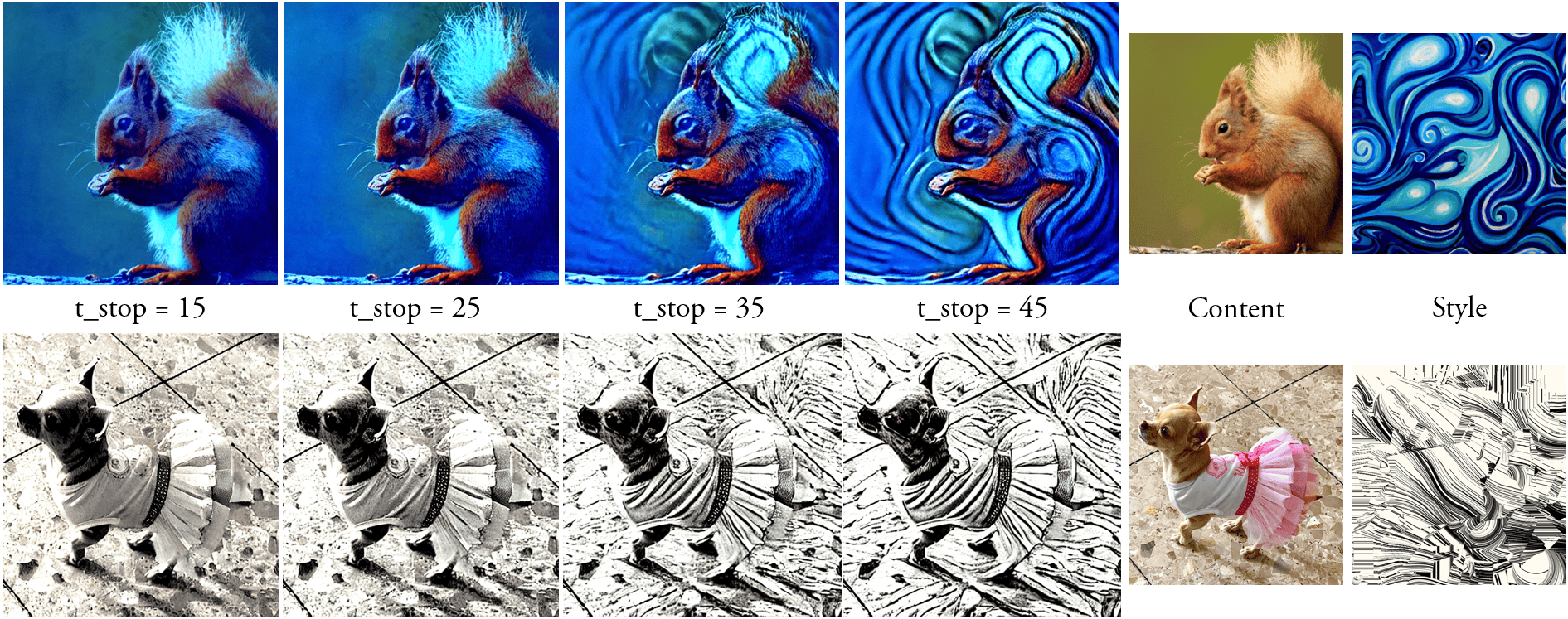}
    \squeezeup
    \caption{Controlling the stylization strength by varying the stopping timestep at which to apply attention modifications. This inference-time control vector affects the content deformity less than varying noise injection timesteps.}
    \label{fig:ending_attn}
    \squeezeup
    \squeezeup
    \squeezeup
    \squeezeup
\end{figure}

\section{Experiments and Evaluation}

Neural style transfer using diffusion models is a nascent sub-field of research. As such, very few works study this new direction, much less via prompt-less techniques. Despite not being a strictly NST model, PARASOL \cite{parasol} is currently the only suitable method we can baseline against. We additionally compare against three recent "traditional" NST techniques, NNST \cite{nnst}, NeAT \cite{neat}, and CAST \cite{cast}. These techniques have focused on texture-based style transfer, and as such, their stylized outputs contain a much better match between the style and stylized images' textures. This is reflected in metrics such as SIFID \cite{singan}, used in NST literature so far that precisely measure such correlations.

The unrolled approach of training diffusion models does incur a high computation cost. Our technique can train over an LDM at 512px resolution on a GPU with 48GB VRAM at batch size 1. We use gradient accumulation 8 to raise the effective batch size to 8. Inference at 512px fits on 24GB VRAM. We train our model for 3 weeks on a single A100. Like NeAT \cite{neat}, we use the BBST-4M dataset they introduce, due to its great variety of style data, covering not just fine-art imagery as more commonly found in other datasets. Due to our method and NeAT having been trained using BBST-4M, we aim to use a test set with no overlap with training data. We use the test set from ALADIN-NST \cite{aladinnst}, which was collected as a test set not overlapping with previous datasets such as BBST-4M. The test set contains 100 content and 400 style images, resulting in 40,000 stylized images. We collect quantitative metrics in Table 1, measuring SIFID \cite{singan} and Chamfer for style and color consistency with the style image respectively, and LPIPS \cite{lpips} for structure consistency with the content. Due to long-running generation times for our method and those of multiple baselines, we randomly sub-sample and use 5,000 images. 

\begin{minipage}[c]{0.385\textwidth}
    \centering
    \label{tab:quantitative_metrics}
    \captionof{table}{Quantitative metrics. \newline Lower is better. $\downarrow$}
    \begin{adjustbox}{width=0.9\textwidth}
        \begin{tabular}{lccc}
            \toprule
            Model & LPIPS $\downarrow$ & SIFID $\downarrow$ & Chamfer $\downarrow$ \\
            \midrule
            NeAT \cite{neat} & 0.624 & 0.880 & 24.970 \\  
            CAST \cite{cast} & 0.632 & 1.520 & 43.864 \\  
            NNST \cite{nnst} & 0.633 & 2.007 & 53.328 \\  
            PARASOL \cite{parasol} & 0.716 & 3.297 & 105.371 \\ 
            \midrule
            DIFF-NST (Ours) & 0.656 & 2.026 & 45.777 \\ 
            \bottomrule    
        \end{tabular}
    \end{adjustbox}
\end{minipage}
\begin{minipage}[c]{0.629\textwidth}
    \centering
    \label{tab:results_mturk}
    \captionof{table}{User studies for our model, for individual ratings (out of 5), and 5-way preferences (\%). Higher is better. $\uparrow$}
    \begin{adjustbox}{width=1\textwidth}
        \begin{tabular}{lcc|cc}
            \toprule
            Model & Content Rating $\uparrow$ & Style Rating $\uparrow$ & Content Preference $\uparrow$ & Style Preference $\uparrow$ \\
            \midrule
            NeAT \cite{neat} & 3.271 & 2.952 & 32.222 & 26.000 \\
            CAST \cite{cast} & 3.031 & 2.863 & 16.756 & 16.133 \\
            NNST \cite{nnst} & 2.937 & 2.712 & 21.200 & 17.778 \\
            PARASOL \cite{parasol} & 2.301 & 2.257 & 12.400 & 9.556 \\
            \midrule
            DIFF-NST (Ours) & 2.751 & 2.973 & 17.422 & 30.533 \\            
            \bottomrule
        \end{tabular}
    \end{adjustbox}
\end{minipage}

We present a qualitative random sample of stylizations in Fig \ref{fig:extra_deform} and the supplementary materials. We visualize stylizations using our method, the closest technically related work PARASOL \cite{parasol}, and some traditional NST techniques. 

The most impactful ablation to report on is experimenting with the style embedding used alongside the style attention values. We show some comparative examples in Fig \ref{fig:horror_faces}, having tested the regular ALADIN-ViT style embedding and the more disentangled ALADIN-NST variant. The ViT variant introduces some content features from the style image into the stylized image when these features have strong activations - most commonly occurring with faces. Though rare, we mitigate this issue using a fully disentangled style embedding, ALADIN-NST. 

\begin{figure}
    \centering
    \includegraphics[width=\linewidth]{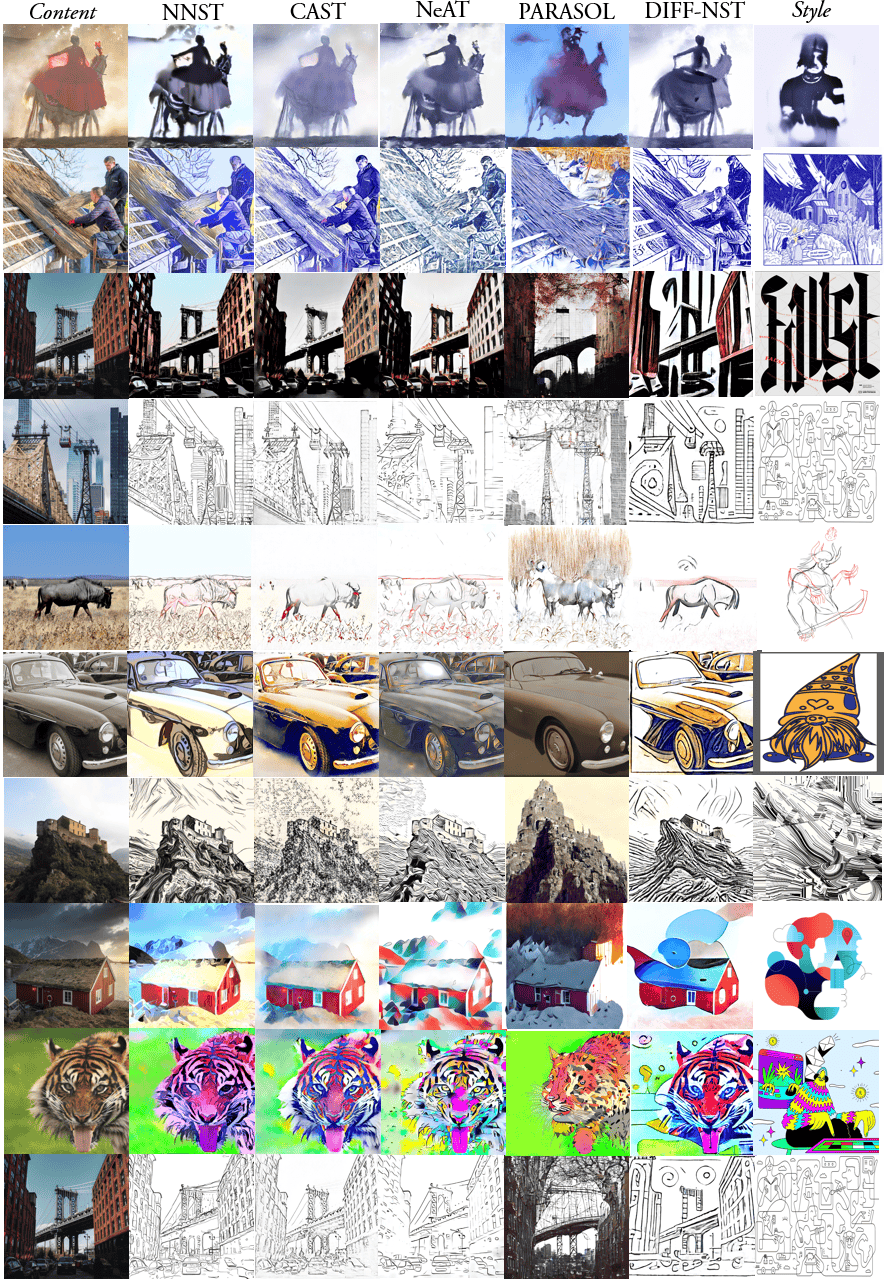}
    \caption{Deformable style transfer, comparing to NNST \cite{nnst}, CAST \cite{cast}, NeAT \cite{neat}, and PARASOL \cite{parasol}. All our figures are generated using images from the ALADIN-NST test set, which were not seen during training. More in the supplementary materials.}
    \label{fig:extra_deform}
\end{figure}

\subsection{User studies}

We undertake a pair of user studies to gauge real life human preference amongst our method and the baselines. First, we carry out an individual rating exercise, measuring the content fidelity between the content image and the stylized image, and separately measuring the style consistence compared to the style image. Second, we carry out a 5-way comparison, where we ask workers to select their best preference from randomly shuffled samples. We bin the ratings in the individual exercise to five levels, and we explicitly instruct what each rating level should represent. We include the definitions in the supplementary material. We randomly sub-sample 750 stylized samples from the test set and compare our method against each baseline on Amazon Mechanical Turk (AMT). We collect and average our responses over 5 different workers for each comparison, and show our results in Table 2.

The results indicate that workers are scoring our DIFF-NST method low on the content information, in both the ratings and preference studies. This is a positive result, as it highlights our technique's more substantial content deformation. The only model which scored lower is PARASOL. However, as seen in our visual comparison figures, PARASOL tends to make significant conceptual changes to the depicted content. It is not so much a technique for style transfer as it is for style-inspired re-generation of similar semantic content. The results for our style-focused experiments indicate that workers prefer our method to baselines in both individual ratings and 5-way preference studies, which signifies a successful transfer of style while still deforming the content.

\subsection{Inference controls}

One key strength of our diffusion-based NST method is control over the structural deformity in the represented content concerning the style image. The reference content information is injected into the diffusion process by applying noises at each time step, pre-extracted from the content image inversion. With diffusion models, the early time steps strongly affect the significant structural components of the image, whereas the later timesteps affect lower level textural information. Therefore, by varying the starting timestep at which these pre-extracted content noises are applied, we can adjust, at inference time, how much the style should deform the content structure. This effect is difficult to evaluate quantitatively, but we show two examples in Fig \ref{fig:starting_noises}.

An alternative vector of inference-time control is varying the diffusion timesteps in which our method's attention replacement happens. By stopping at earlier timesteps, less style information is injected into the diffusion process, reducing the stylization strength. Unlike reducing content noise injection, this approach maintains the content structure better and more directly targets the style properties instead of structure. We show examples of this second approach in Fig \ref{fig:ending_attn}, using the same example images as in Fig \ref{fig:starting_noises} for clarity.

\section{Limitations and Conclusions}
\squeezeup

One limiting factor of our approach is that textures are not matched to the style image with as much detail and fidelity as traditional NST approaches. This can, however, be alleviated by introducing a conventional NST approach into the pipeline as a post-processing step.

Though rare, due to the one-to-one mapping between the content and style attention values, some structure from some style images sometimes creeps into the stylized image. We can report negative results experimenting with Neural Neighbours \cite{nnst} in attention space, which resolved this issue, but only at the cost of worse overall stylization quality. This is an area of potential future improvement.

One of the principal challenges with our method has been computation due to the unrolled nature of the reverse diffusion process during training. Future work can explore the adaptation of the style training objective to the latent space instead of pixel space, enabling non-unrolled training.

\section{Broader Impact}

Neural techniques for artistic image editing and generation offer new tools and capabilities for skilled artists to take their work further than before. However, this does make the field easier to enter as a novice. As such, existing novice-level artists may find more competition in this space, reducing work opportunities. As digital art emerged, it offered new capabilities to artists with new tools at the detriment of some artists using physical mediums. Neural techniques can similarly open up new genres of art while reducing some opportunities for some existing digital artists.

\medskip
\small

\bibliographystyle{plainnat}
\bibliography{references.bib}

\appendix

\section{Prompt-to-prompt Analysis}

The base content captions partially generated using ChatGPT for the prompt-to-prompt analysis experiments are:

\begin{enumerate}
    \item A squirrel eating a burger
    \item A hamster on a skateboard
    \item A toy next to a flower
    \item A car driving down the road
    \item A giraffe in a chair
    \item A bear wearing sunglasses
    \item An octopus in a space suit
    \item A hedgehog getting a haircut
    \item A sloth running a marathon
    \item A cat posing like napoleon
    \item A dog with a beard, smoking a cigar
    \item A bee flying underwater next to fish
    \item A fish with a hat, playing a guitar
    \item A bird with a bowtie, playing a saxophone
    \item A turtle with a top hat, playing a piano
    \item A frog with a cowboy hat, playing a banjo
    \item A mouse with a sombrero, playing a trumpet
    \item A snake with a beret, playing a violin
    \item A rabbit with a fedora, playing a cello
    \item A squirrel with a baseball cap, playing a drum
\end{enumerate}

\begin{figure}
    \centering
    \includegraphics[width=0.7\linewidth]{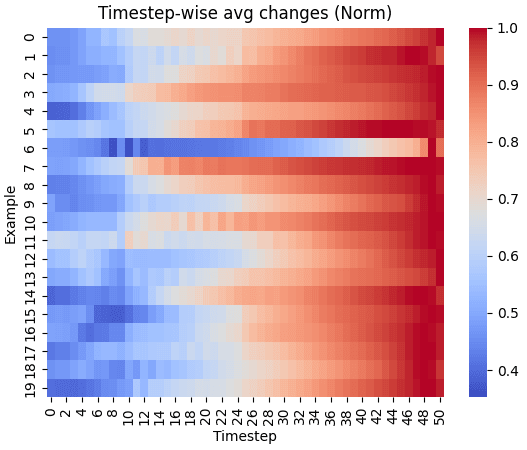}
    \caption{The averaged normalized difference in attention values between reference content and 10 stylized content images in the prompt-based prompt-to-prompt model. The higher difference values (in red) in the later timesteps visualize the effect that earlier timesteps affect coarser structural details, whereas later timesteps affect lower level textural details.}
    \label{fig:p2p_timestep}
\end{figure}

The style modifiers are:

\begin{enumerate}
    \item A van gogh painting of
    \item A graphite sketch of
    \item A neon colourful pastel of
    \item A minimal flat vector art illustration of
    \item A watercolour painting of
    \item A psychedelic inverted painting of
    \item A pop-art comic book panel of
    \item A neoclassical painting of
    \item A cubist abstract painting of
    \item A surreal dark horror painting of
\end{enumerate}

We visualize results from the preliminary prompt-to-prompt analysis experiments, in Fig \ref{fig:p2p_interp}. The figure shows the first content prompt for the base content, with the subsequent rows interpolating towards style-modified prompts using style prompt modifiers 1, 4, 2, and 8. Although not directly relevant to our study, it was also interesting to note that the stylization strength could be pushed beyond the default strength by pushing the interpolation into over-drive, similar to the technique presented in NeAT \cite{neat}.

\begin{figure}
    \centering
    \includegraphics[width=\linewidth]{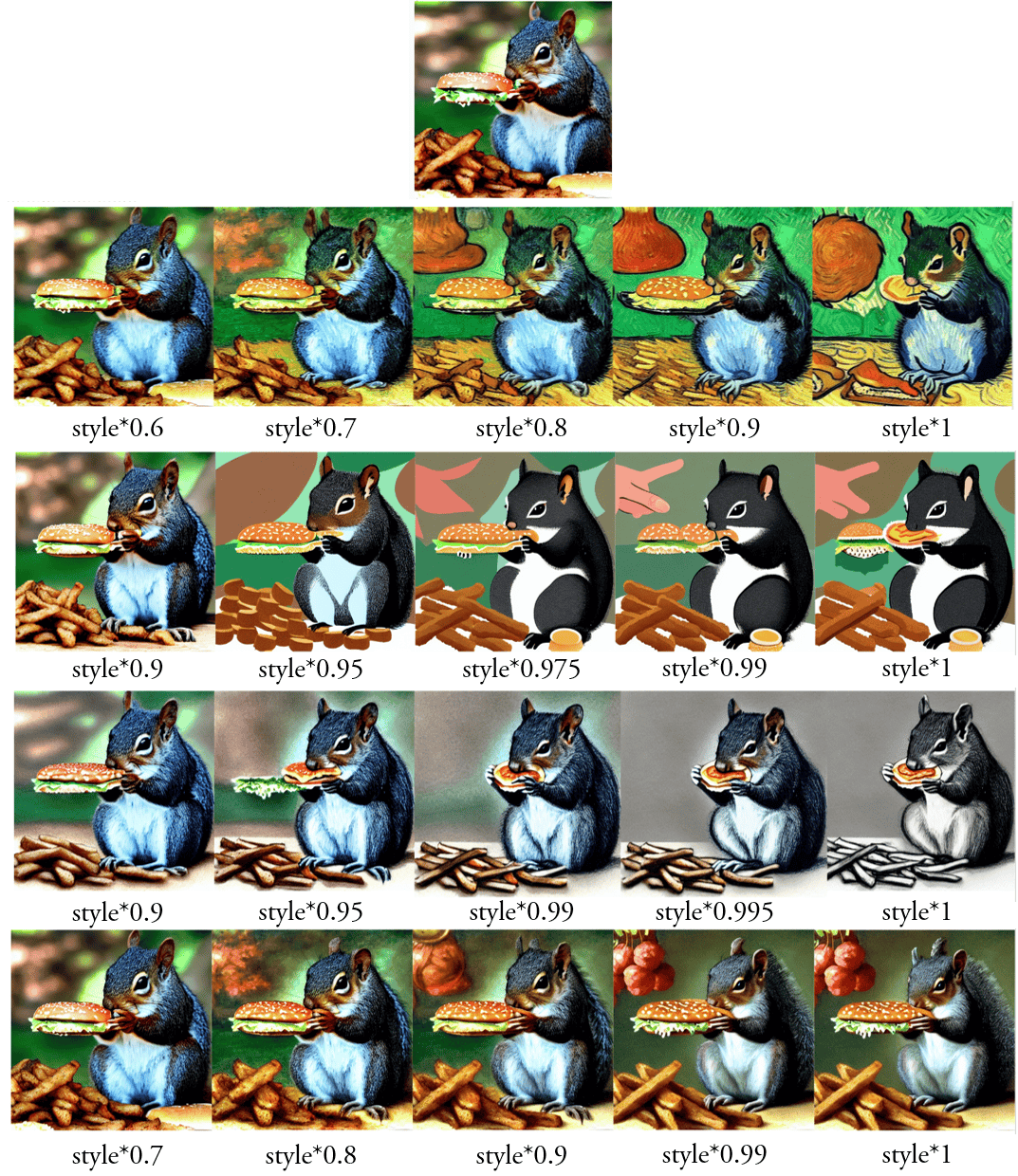}
    \caption{Visualization of stylization interpolation using prompt-to-prompt, changing only the attention V values. The stylization strength displayed represents the interpolation strength between the content and style attention values.}
    \label{fig:p2p_interp}
\end{figure}

\begin{figure}
    \centering
    \includegraphics[width=\linewidth]{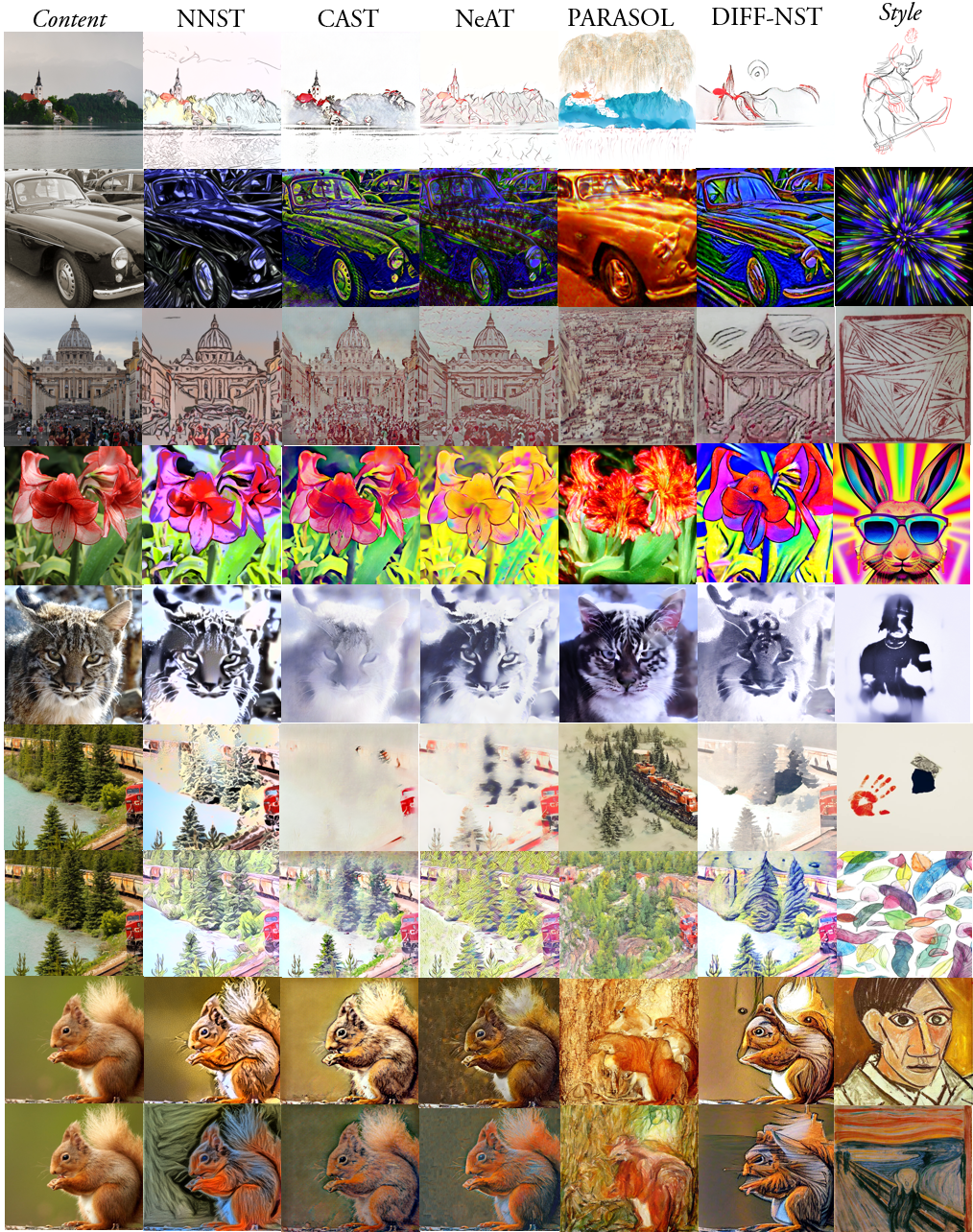}
    \caption{Additional deformable style transfer comparisons}
    \label{fig:extra_deform}
\end{figure}

\section{Additional details on user studies}

We carried out two user studies: an individual rating exercise with defined rating levels, and a 5-way preference comparative exercise. For each, we executed the experiments once for the content, and once for the style.

Our content-focused rating exercise asks the following question: "A photo has been re-generated with a different style. Please rate the structure details of the new image, 1 to 5 as follows:", where we next define the expected judgement criteria for each rating level as follows:

\begin{enumerate}
    \item The structure is different
    \item The structure slightly resembles the photo 
    \item The structure mostly resembles the photo
    \item The structure is the same
    \item The structure is the same, including small details
\end{enumerate}

Our style focused rating exercise asks the following question: "A photo has been transformed into the style of the artwork. Please rate the quality of the style, 1 to 5 as follows:", where the rating definitions are:

\begin{enumerate}
    \item The style is not recognisable
    \item The style is recognisable
    \item The colours match
    \item The textures match
    \item The shapes match
\end{enumerate}

The 5-way comparative study presents the following question for the content-focused experiment: "A photo has been re-generated with a different style in 5 ways. Please select the highest quality reconstruction of the photo's structure details", and the following for the style-focused experiment: "A photo has been re-generated with a different style in 5 ways. Please select the most similar artistic style to the artwork"

The workers were fairly compensated. We used 5 different workers for each stylized image, for each question.

\end{document}